\documentclass[10pt, a4paper]{article}
\usepackage{booktabs}
\usepackage{amsmath}
\usepackage{graphicx}
\usepackage[final]{lrec2026}
\title{TeamHerald@CHIPSAL 2026: Hate Speech Detection and Sentiment Analysis of Nepali Memes using Transformer-based Architectures and Ensemble Learning}

\name{Ashish Acharya$^*$, Anish Khatiwada$^*$, Rohit Khadka, Pragya Aryal}

\address{
  Herald College Kathmandu, Kathmandu, Nepal \\
  {\normalsize \{ashish.acharya048, anishkhatioda, rohitkhadka153, pragyaaryal18\}@gmail.com} \\
  \vspace{0.2cm}
  \small $^*$These authors contributed equally to this work.
}

\abstract{
The analysis of internet memes in the Nepali language is complicated by frequent code-mixing and a lack of established baseline resources. While memes inherently combine visual and textual elements, this study focuses on a text-centric approach by extracting embedded text using an OCR layer and modeling it with Transformer-based architectures. We evaluate six distinct models and investigate the comparative effectiveness of Hard and Soft Voting ensemble strategies across two tasks: binary hate speech detection and three-class sentiment analysis. Experimental results show that a standalone decoder-only model achieved the highest performance for binary classification, whereas the Soft Voting ensemble performed best for the multi-class sentiment task, yielding a 15.8\% relative improvement in Macro F1-score over the strongest standalone baseline. These findings suggest that ensemble strategies behave differently across binary and multi-class tasks, highlighting the importance of selecting aggregation methods suited to the classification objective.
\\ \newline  }

\begin{document}
\maketitleabstract
\thispagestyle{empty}
\section{Introduction}
Internet memes have evolved from simple units of humor into sophisticated instruments for socio-political discourse, cultural expression, and, increasingly, the dissemination of harmful content. Unlike traditional text-only social media posts, memes are inherently multimodal, with meaning emerging from the interaction between visual imagery and embedded text \cite{kiela2020hateful}. This combination often creates compositional effects, where the sentiment of the text can be inverted or intensified by the accompanying image, rendering unimodal detection systems largely ineffective \cite{pramanick2022momenta}.

For low-resource languages such as Nepali, these challenges are compounded. The linguistic landscape features frequent code-mixing, including Devanagari and Romanized Nepali alongside English, which complicates standard tokenization and semantic analysis \cite{rauniyar2023nepali}. In addition, there is a scarcity of high-quality, annotated multimodal datasets that capture the cultural and political nuances of Nepali digital content. While significant progress has been made in English-centric meme analysis, such models often fail to generalize to South Asian contexts, where hate speech and sentiment are closely tied to local events, ethnic identities, and region-specific political satire.

To address these limitations, this paper evaluates a robust text-centric pipeline for Nepali meme analysis. An Optical Character Recognition (OCR) layer extracts textual content from raw images, simulating a real-world scenario in which ground-truth text is unavailable. We then benchmark six Transformer-based architectures, including models pre-trained on Nepali and cross-lingual corpora, to identify the most effective base learners for this domain.

The main contribution of this work is a systematic evaluation of ensemble strategies within a multimodal framework. By applying both Hard (majority-based) and Soft (probability-based) voting across binary and multi-class tasks, we examine how ensemble logic interacts with task complexity. Our analysis shows that Hard Voting stabilizes binary hate speech detection by mitigating model-specific noise, while Soft Voting is more effective for multi-class sentiment classification. This indicates that preserving the probabilistic confidence of underlying models is beneficial for resolving subtle linguistic ambiguities in multi-class social media discourse.
\section{Related Work}
\label{sec:related}
Significant advances have been made in hate speech detection and sentiment analysis, particularly for high-resource languages such as English. However, low-resource languages, including Nepali, remain underrepresented in both dataset availability and model development \citep{parihar2021hate}. The challenge is further amplified in the context of multimodal content such as memes, where meaning emerges from the interaction between textual and visual modalities. 
\subsection{Hate Speech and Sentiment Analysis in Low-Resource Languages}

Hate speech and sentiment analysis in low-resource languages has gained increasing attention due to the rapid expansion of South Asian social media content. However, compared to English, languages such as Nepali, Hindi, Bangla, and Hinglish present significant challenges due to limited annotated datasets, linguistic diversity, and frequent code-mixing.

Early sentiment analysis work in Nepali relied on traditional machine learning models such as Support Vector Machines (SVM) and Naive Bayes with hand-crafted features \citep{sitoula2021nepali}. These approaches demonstrated limited generalization capability. The emergence of deep learning enabled the use of Convolutional Neural Networks (CNNs) and Long Short-Term Memory (LSTM) networks, which improved performance through contextual embeddings. \citep{joshi2019deep}

For Bangla and Hindi, several studies have investigated abusive and hate speech detection in social media, emphasizing issues such as informal spelling, transliteration, and cultural nuance \citep{Das2023BanglaAbuseMemeAD}. Transformer-based models, particularly multilingual and monolingual pretrained architectures, have consistently demonstrated improved performance in these settings \citep{rathore2025hinglish, das-etal-2022-hatecheckhin}. 

However, recent findings suggest that domain-relevant non-hateful pretraining data may yield performance improvements comparable to explicitly hateful pretraining, challenging assumptions about task-specific pretraining strategies \citep{rottger2022undermining}. Despite progress, data scarcity and domain adaptation remain core challenges in low-resource setups.

\subsection{Multimodal Hate and Meme Analysis}

With memes becoming a dominant medium for political expression and social commentary, hate speech detection has expanded beyond text-only analysis to multimodal settings. Memes often encode meaning through the interaction of visual and textual components, making unimodal models insufficient \citep{kiela2022creative}.

Previous multimodal hate detection research shows that models integrating both visual and textual features outperform text-only approaches \citep{pramanick2022memeclip}. In many cases, hateful or sarcastic intent is not explicit in either modality alone but emerges from their combination \citep{bohra2020codemixed}. Election-related memes, particularly in South Asian contexts, have been shown to propagate hate and misinformation, requiring culturally informed modeling approaches \citep{article}.

Prompt-based and contrastive learning methods have recently gained attention for modeling implicit cross-modal relationships without requiring extensive labeled datasets \citep{ranasinghe2021multilingual}. CLIP-based architectures and adapter-based fine-tuning strategies further enhance multimodal meme classification, though detecting creative, humorous, or sarcastic hateful memes remains an open challenge, especially in multilingual and low-resource environments \citep{pramanick2022memeclip, kiela2022creative}.

\subsection{Datasets and Annotation Challenges}

A major bottleneck in low-resource hate speech and sentiment analysis is the scarcity of high-quality annotated datasets. Creating datasets for languages such as Nepali, Hindi, and Bangla requires linguistic expertise and cultural understanding, particularly when addressing political or election-related discourse \citep{article}.

Multimodal meme datasets introduce additional complexity, as annotators must interpret visual context, textual meaning, humor, sarcasm, and socio-political nuance simultaneously \citep{kiela2022creative, pramanick2022memeclip}. Many existing South Asian hate speech datasets remain text-only and relatively small in scale \citep{Das2023BanglaAbuseMemeAD, rottger2022undermining}.

Code-mixed datasets further complicate annotation due to ambiguous language boundaries, inconsistent spelling, and annotator disagreement \citep{rathore2025hinglish}. Studies frequently report moderate inter-annotator agreement for subjective labels such as hate and sarcasm \citep{ranasinghe2021multilingual}. These challenges highlight the need for clearer annotation protocols and scalable dataset creation strategies in low-resource multimodal research.

\subsection{Modeling Approaches}

Modeling strategies for hate speech and sentiment classification have evolved from traditional bag-of-words and TF-IDF approaches to deep neural architectures \citep{sitoula2021nepali, Das2023BanglaAbuseMemeAD}. While traditional models are computationally efficient, they struggle to capture semantic and contextual nuance in informal social media text.

Recurrent neural networks (RNNs), LSTMs, and GRUs improved contextual modeling capabilities, particularly in Nepali and Hindi sentiment analysis \citep{joshi2019deep}. Transformer-based architectures further advanced the field through large-scale pretraining and cross-lingual transfer learning \citep{rathore2025hinglish, das-etal-2022-hatecheckhin}. 

In multimodal scenarios, vision-language transformer models such as CLIP have shown strong performance by learning joint image-text representations \citep{pramanick2022memeclip}. Prompt-based learning and lightweight adaptation methods offer promising directions for low-resource meme classification tasks \citep{ranasinghe2021multilingual}. Nevertheless, interpretability and cultural sensitivity of large transformer models in low-resource contexts remain underexplored.

Despite substantial progress, several gaps persist. First, limited work simultaneously addresses low-resource languages, code-mixing, and multimodal meme analysis within a unified framework. Second, Nepali multimodal meme datasets remain scarce, restricting comprehensive evaluation. Third, although transformer and CLIP-based models demonstrate strong empirical performance, their interpretability and socio-cultural alignment in South Asian contexts require further investigation. Finally, scalable annotation strategies and weakly supervised or prompt-based methods are necessary to reduce reliance on large labeled datasets.

Addressing these gaps is critical for developing robust hate speech and sentiment classification systems for multilingual, low-resource, and multimodal social media environments.

\section{Datasets and Tasks Description}
The shared task consists of two different subtasks: Subtask A aims to detect hate speech in Nepali meme images, Subtask B is related to classifying the sentiment from Nepali meme images. For all of the mentioned subtasks, datasets were provided by the organizers, which were initially created and curated by different papers \citep{thapa2026sharedtask, sarves2026chipsal}.

\subsection{Sub-Task A}
 It involves binary classification to distinguish between images classified as HATE (labeled as 1) and NO-HATE (labeled as 0) \citep{thapa2025multimodal, thapa2025cross, bhandari2023crisishatemm}. For the training purposes of this sub-task, a total of 1068 datasets were provided, out of which 720 were labeled as Hate and 348 were labeled as No Hate. The other 133 datasets were provided for validation, and 134 for testing purposes.

\subsection{Sub-Task B}
 It involves multi-class classification to distinguish between images classified as Negative (labeled as 0), Neutral (labeled as 1), and Positive (labeled as 2) \citep{thapa2025multimodal,thapa2025cross,bhandari2023crisishatemm}. For the training purpose of this sub-task, a total of 1061 datasets were provided, out of which 473 were labeled as Negative, 341 were labeled as Neutral and 247 were labeled as Positive. The other 133 data sets were for validation, and 133 for testing purposes.

\section{Methodology}
\label{sec:methodology}
Our approach was designed to navigate the inherent multimodality of memes, where visual and textual cues are often inextricably linked. We developed a pipeline that first extracts embedded text via EasyOCR and subsequently processes these strings through a Transformer-based ensemble. While ground-truth text labels were available during the training phase, we intentionally restricted the testing phase to raw images. This design choice was made to simulate a real-world deployment scenario where OCR noise and extraction failures are inevitable \citep{rauniyar2023nepali}.

\subsection{Dataset Balancing}
A major challenge in our data pipeline was class imbalance, which can bias models toward majority labels and negatively impact Macro F1-score. To address this, we applied random oversampling to the minority classes to obtain a more balanced training distribution. Specifically, samples were resampled with replacement to reach $n = 1000$ instances per class for Subtask B and $n = 1500$ for Subtask A.

These target sizes were selected based on preliminary experiments, where increasing the representation of minority classes led to improved validation performance, particularly for Subtask A. This approach helps ensure that all classes contribute more evenly to the loss function during training, reducing bias toward majority classes.

\subsection{Model Selection}
To capture code-mixed and low-resource linguistic features, we fine-tuned six Transformer-based architectures from Hugging Face, each selected for specific advantages:
\begin{itemize}
\item \textbf{NepaliBERT}: BERT-base model specialized for native Nepali semantics, providing a robust baseline for formal and semi-formal Devanagari text.  

\item \textbf{DistilBERT-Base-Nepali}: Knowledge distilled, computationally efficient variant for Nepali text, included to evaluate performance trade-offs in resource-constrained environments.  

\item \textbf{XLM-RoBERTa-Base}: Multilingual model capable of handling complex code-mixed content by leveraging shared representations across 100 languages.  

\item \textbf{RoBERTa-Hindi}: Used for cross-lingual transfer learning; shared Devanagari script and lexical overlap between Hindi and Nepali may enhance feature representation for low-resource tasks.  

\item \textbf{distilgpt2-nepali}: Decoder-only GPT-2 model with continual pre-training on 13 million Nepali sequences. This model was evaluated only as a standalone baseline and was not included in the ensemble.  
\end{itemize}

\subsection{Experimental Setup}
All models were fine-tuned using the parameters detailed in Table~\ref{tab:params}. We implemented \textbf{Early Stopping} with a patience of 3 epochs and a threshold of 0.01 to monitor validation loss.

\begin{table}[h]
\centering
\caption{Fine-Tuning Hyperparameters}
\label{tab:params}
\begin{tabular}{@{}ll@{}}
\toprule
\textbf{Hyperparameter} & \textbf{Value} \\ \midrule
Epochs                  & 10             \\
Batch Size              & 16             \\
Learning Rate           & $2 \times 10^{-5}$ \\
Weight Decay            & 0.01           \\
Early Stopping Patience & 3              \\
Logging Steps           & 50             \\ \bottomrule
\end{tabular}
\end{table}

\subsection{Evaluation Framework: Individual and Ensemble Testing}
Following the evaluation of the six standalone models, we implemented an ensemble layer to aggregate their predictions. This was conducted using two distinct voting strategies applied across both subtasks to identify the most robust method for final classification.

\subsubsection{Hard Voting}
In the Hard Voting setup, the final prediction $\hat{y}$ is derived from a majority consensus of the five base learners. In this configuration, each model contributes a single discrete label, and the most frequent class is selected as the final output. This method treats all models as equal contributors regardless of their individual confidence scores.

\subsubsection{Soft Voting}
We concurrently tested a Soft Voting strategy, which aggregates the predicted class probabilities (softmax outputs) rather than discrete labels. The final predicted class $\hat{y}$ is calculated as:$$\hat{y} = \arg \max_{i} \sum_{j=1}^{5} w_j p_{ij}$$where $p_{ij}$ represents the probability assigned to class $i$ by model $j$. This allows the ensemble to account for the relative confidence of each base learner across the 2-class (Subtask A) and 3-class (Subtask B) problems.

\section{Results and Analysis}
\label{sec:results}
The quantitative performance across both subtasks, summarized in Table~\ref{tab:results_final}, highlights architectural differences in low-resource multimodal NLP settings. In accordance with the official shared task evaluation criteria, Macro F1-score is reported as the primary metric.

In Subtask A (Hate Speech), the standalone Sakonii/distilgpt2-nepali model achieved the highest performance ($F1=0.6550$). This result may indicate that domain-specific continual pre-training contributed to improved binary discrimination between offensive and non-offensive content. Among the encoder-based models, RoBERTa-Hindi and NepaliBERT achieved identical evaluation metrics across all reported measures; however, their predictions differ at the instance level, suggesting similar class-wise error distributions. Although the ensemble approaches (Hard and Soft Voting) remained competitive, they did not surpass the GPT-2 baseline. This suggests that for binary classification tasks, a well-adapted single model can perform comparably to or better than aggregation-based methods.

In contrast, Subtask B (Sentiment Classification) revealed greater benefits from model aggregation under a three-class setting. The Soft Voting ensemble achieved the highest Macro F1-score (0.5518), representing a 15.8\% relative improvement over the strongest standalone baseline (RoBERTa-Hindi). This pattern suggests that probabilistic aggregation may better capture uncertainty in multi-class sentiment classification compared to majority-based voting. Overall, these findings indicate that in low-resource settings, performance gains may depend not only on model selection but also on the aggregation strategy employed.

\begin{table*}[t]
\centering
\small
\caption{Experimental Results: Comparative Performance of Standalone Base Learners and Ensemble Strategies across Subtasks A and B.}
\label{tab:results_final}
\begin{tabular}{@{}llcccc@{}}
\toprule
\textbf{Task} & \textbf{Model/Strategy} & \textbf{F1 Macro} & \textbf{Accuracy} & \textbf{Precision} & \textbf{Recall} \\ \midrule
\textbf{Subtask A} & Twitter-XLM-RoBERTa-Base & 0.5830 & 0.6343 & 0.5836 & 0.5826 \\
(Hate Speech) & DistilBERT-Base-Nepali & 0.5318 & 0.5970 & 0.5332 & 0.5316 \\
 & RoBERTa-Hindi & 0.5878 & 0.6343 & 0.5874 & 0.5884 \\
 & NepaliBERT & 0.5878 & 0.6343 & 0.5874 & 0.5884 \\
 & XLM-RoBERTa-Base & 0.5387 & 0.6343 & 0.5545 & 0.5419 \\
 & \textbf{Sakonii/distilgpt2-nepali} & \textbf{0.6550} & \textbf{0.7090} & \textbf{0.6652} & \textbf{0.6497} \\ 
 \cmidrule(l){2-6} 
 & Hard Voting (Ensemble) & 0.6422 & 0.6940 & 0.6482 & 0.6386 \\
 & Soft Voting (Ensemble) & 0.6310 & 0.6866 & 0.6382 & 0.6273 \\ \midrule
\textbf{Subtask B} & Twitter-XLM-RoBERTa-Base & 0.4110 & 0.4286 & 0.4456 & 0.4333 \\
(Sentiment) & DistilBERT-Base-Nepali & 0.4341 & 0.4586 & 0.4341 & 0.4352 \\
 & RoBERTa-Hindi & 0.4764 & \textbf{0.5789} & \textbf{0.6481} & 0.4918 \\
 & NepaliBERT & 0.3945 & 0.3985 & 0.3955 & 0.4083 \\
 & XLM-RoBERTa-Base & 0.4484 & 0.4511 & 0.4484 & 0.4623 \\
 & Sakonii/distilgpt2-nepali & 0.4108 & 0.4436 & 0.4222 & 0.4077 \\ 
 \cmidrule(l){2-6} 
 & Hard Voting (Ensemble) & 0.4622 & 0.4887 & 0.5082 & 0.4635 \\
 & \textbf{Soft Voting (Ensemble)} & \textbf{0.5518} & 0.5639 & 0.5659 & \textbf{0.5452} \\ \bottomrule
\end{tabular}
\end{table*}

\section{Limitations}
\label{sec:limitations}
While the proposed approach demonstrates performance gains for Nepali meme classification, several limitations should be considered when interpreting these findings.

First, the evaluation relies on a text-centric pipeline dependent on the EasyOCR extraction layer. Because the classification models operate solely on extracted text without incorporating visual features, overall performance is sensitive to OCR quality. A qualitative inspection of OCR outputs revealed frequent errors in stylized fonts, low-resolution text, and visually complex layouts, often resulting in incomplete or distorted tokens. These issues may lead to error propagation, where downstream classification is performed on noisy or partially incorrect inputs.

Second, the linguistic complexity of the dataset remains a persistent challenge. The prevalence of code-mixing (Nepali, English, and Romanized Nepali), along with colloquial expressions, may not be fully captured by existing pre-training corpora. This limitation is particularly evident in the three-class sentiment task, where subtle contextual and semantic distinctions are required to differentiate between neutral and polarized classes.

Finally, the use of random oversampling to balance class distributions creates a controlled training setup that does not reflect the naturally skewed distribution of hate speech in real-world social media. As a result, the reported performance should be interpreted within the context of balanced experimental conditions rather than as a direct estimate of real-world deployment performance.

\section{Conclusion}
\label{sec:conclusion}

We evaluated several Transformer architectures and ensemble strategies for hate speech and sentiment analysis in the low-resource Nepali language. The experimental results suggest differing strengths between standalone and ensemble architectures across tasks. While a standalone decoder-only model (Sakonii/distilgpt2-nepali) achieved the highest performance in binary hate speech detection, the Soft Voting ensemble achieved the strongest performance in the three-class sentiment task. Specifically, the 15.8\% relative improvement in F1-Macro for Subtask B suggests that probabilistic aggregation may be beneficial for multi-class sentiment classification.

These findings indicate that performance in this task varies with class cardinality and aggregation strategy. While memes are inherently multimodal, this work adopts a text-centric approach based on OCR-extracted content, which may limit performance in cases where meaning is conveyed primarily through visual context. Future work will explore the integration of visual-semantic features, including models such as CLIP or Vision Transformers (ViT), to better capture image-text interactions. Additionally, the application of parameter-efficient fine-tuning techniques remains a promising direction to investigate the balance between the computational requirements of ensemble-based inference and classification accuracy.
\section{Bibliographical References}\label{sec:reference}

\bibliographystyle{lrec2026-natbib}
\bibliography{lrec2026-example,literature_review}

\end{document}